\RequirePackage{fix-cm}

\documentclass[smallcondensed]{svjour3}     % onecolumn (ditto)

\smartqed  % flush right qed marks, e.g. at end of proof
\usepackage{graphicx}
\usepackage{listings}
\usepackage{verbatim}
\usepackage{graphicx}
\usepackage{url}
\usepackage{fancyhdr}
\usepackage{fancybox,framed}
\usepackage{lastpage}
\usepackage{floatrow}
\usepackage{booktabs,xltabular}
\usepackage{makecell}
\usepackage[table]{xcolor}
\usepackage{amssymb}
\usepackage[toc,page]{appendix}
\usepackage{subcaption}
\usepackage{amsmath}

\usepackage{xcolor}

\newcommand\hl[1]{%
  \bgroup
  \hskip0pt\color{red!80!black}%
  #1%
  \egroup
}

\usepackage[
  separate-uncertainty = true,
  multi-part-units = repeat
]{siunitx}

\usepackage{lipsum}

\DeclareMathOperator*{\argmin}{arg\,min}

\begin{document}

\title{\LARGE \bf Semi-Supervised Adversarial Discriminative Domain Adaptation}

\author{Thai-Vu Nguyen$^{1,2}$, Anh Nguyen$^{3}$, Nghia Le$^{2,4}$, Bac Le$^{1,2}$
}

\institute{$^{1}$ Faculty of Information Technology, University of Science, Ho Chi Minh City, Vietnam \\
%\and
$^{2}$ Vietnam National University, Ho Chi Minh City, Vietnam \\
Email: {\tt{lhbac@fit.hcmus.edu.vn}, \\
\tt{vunguyenthai73@gmail.com}} \\
$^{3}$ Department of Computer Science, University of Liverpool, UK \\
\email{anh.nguyen@liverpool.ac.uk} \\
$^{4}$ University of Information Technology. Ho Chi Minh City, Vietnam \\
Email: \tt{nghialh@uit.edu.vn}
}

\maketitle

\begin{abstract}

Domain adaptation is a potential method to train a powerful deep neural network across various datasets. More precisely, domain adaptation methods train the model on training data and test that model on a completely separate dataset. The adversarial-based adaptation method became popular among other domain adaptation methods. Relying on the idea of GAN, the adversarial-based domain adaptation tries to minimize the distribution between the training and testing dataset based on the adversarial learning process. We observe that the semi-supervised learning approach can combine with the adversarial-based method to solve the domain adaptation problem. In this paper, we propose an improved adversarial domain adaptation method called Semi-Supervised Adversarial Discriminative Domain Adaptation (SADDA), which can outperform other prior domain adaptation methods. We also show that SADDA has a wide range of applications and illustrate the promise of our method for image classification and sentiment classification problems.

\end{abstract}

\section{Introduction}
\label{intro}

Over the past few years, deep neural networks have achieved significant achievements in many applications. One of the major limitations of deep neural networks is the dataset bias or domain shift problems \cite{5376}. These phenomena occur when the model obtains good results on the training dataset; however, showing poor performance on a testing dataset or a real-world sample.

\begin{figure} [htp]
    \includegraphics[width=0.75\columnwidth]{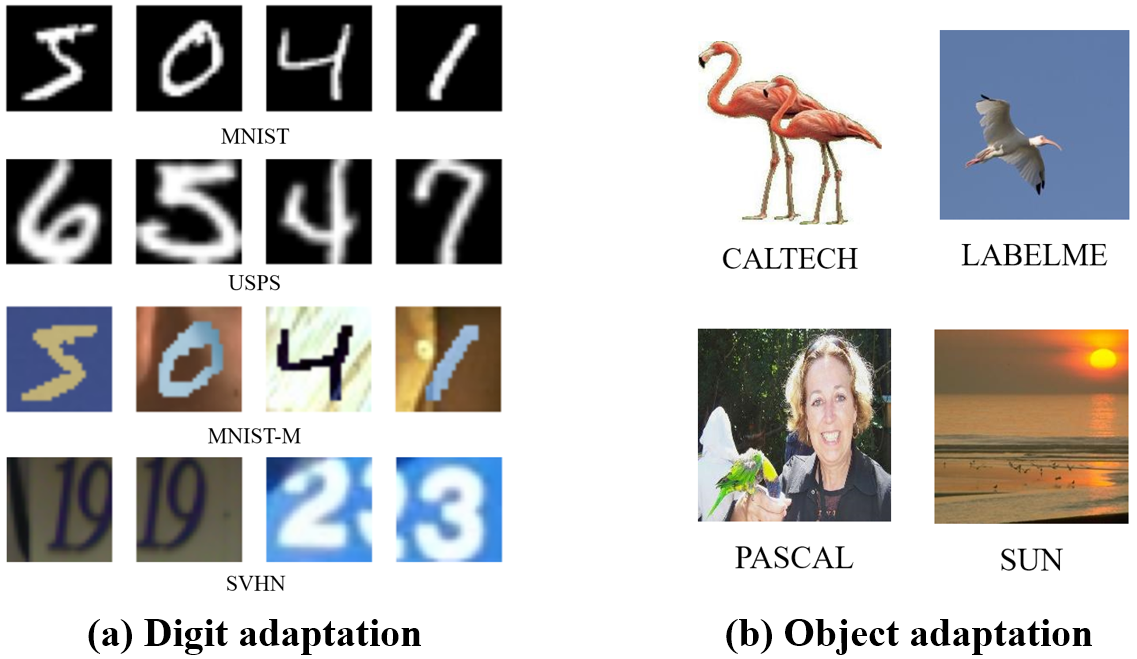}
    \caption{Examples of images from different datasets. (a) Some digit images from MNIST \cite{726791}, USPS \cite{uspsdataset}, MNIST-M \cite{ganin2016domainadversarial}, and SVHN \cite{37648} datasets. (b) Some object images from the "bird" category in CALTECH \cite{1597116}, LABELME \cite{Russell2007LabelMeAD}, PASCAL \cite{pascal-voc-2007}, and SUN \cite{5540221} datasets.}
    \label{fig:samle_images}
\end{figure}

As shown in Figure \ref{fig:samle_images}, because of numerous reasons (illumination, image quality, background), there is always a different distribution between two datasets, which is the main factor reducing the performance of deep neural networks. Even though various research has proved that deep neural networks can learn transferable feature representation over different datasets \cite{Long2015LearningTF,autonomous_nav}, Donahue \emph{et al.} \cite{donahue2013decaf} showed that domain shift still influences the accuracy of the deep neural network when testing these networks in a different dataset.

The solution for the aforementioned problems is domain adaptation techniques \cite{10.1016/j.neucom.2015.03.020,5640675}. The main idea of domain adaptation techniques is to learn how a deep neural network can map the source domain and target domain into a common feature space, which minimize the negative influence of domain shift or dataset bias. 

The adversarial-based adaptation method \cite{NIPS2014_5ca3e9b1,8099799} has become a well-known technique among other domain adaptation methods. Adversarial adaptation includes two networks - an encoder and a discriminator, trained simultaneously with conflicting objectives. The encoder is trained to encode images from the original domain (source domain) and new domain (target domain) such that it puzzles the discriminator. In contrast, the discriminator tries to distinguish between the source and target domain. Recently, Adversarial Discriminative Domain Adaptation (ADDA) by Tzeng \emph{et al.} \cite{8099799} has shown that adversarial adaptation can handle dataset bias and domain shift problems. From there,  we extend the ADDA method to the semi-supervised learning context by obliging the discriminator network to predict class labels.

Semi-supervised learning \cite{semisupervised} is an approach that builds a predictive model with a small labeled dataset and a large unlabeled dataset. The model must learn from the small labeled dataset and somehow exploit the larger unlabeled dataset to classify new samples. In the context of unsupervised domain adaptation tasks, the semi-supervised learning approach needs to take advantage of the labeled source dataset to map to the unlabeled target dataset, thereby correctly classifying the labels of the target dataset. The Semi-Supervised GAN \cite{odena2016semisupervised} is designed to handle the semi-supervised learning tasks and inspired us to develop our model.

In this paper, we present a novel method called Semi-supervised Adversarial Discriminative Domain Adaptation (SADDA), where the discriminator is a multi-class classifier. Instead of only distinguishing between source images and target images (method like ADDA \cite{8099799}), the discriminator learns to distinguish N + 1 classes, where N is the number of classes in the classification task, and the last one uses to distinguish between the source dataset or the target dataset. The discriminator focuses not only on the domain label between two datasets but also on the labeled images from the source dataset, which improves the generalization ability of the discriminator and the encoder as well as the classification accuracy.

To validate the effectiveness of our methodology, we experiment with domain adaptation tasks on digit datasets, including MNIST \cite{726791}, USPS \cite{uspsdataset}, MNIST-M \cite{ganin2016domainadversarial}, and SVHN \cite{37648}. In addition, we also prove the robustness ability of the SADDA method by using t-SNE visualization of the digit datasets, the SADDA method keeps the t-SNE clusters as tight as possible and maximizes the separation between two clusters. We also test its potential with a more sophisticated dataset, by object recognition task with CALTECH \cite{1597116}, LABELME \cite{Russell2007LabelMeAD}, PASCAL \cite{pascal-voc-2007}, and SUN \cite{5540221} datasets. In addition, we evaluate our method for the natural language processing task, with three text datasets including Women's E-Commerce Clothing Reviews \cite{nicapotato_2018}, Coronavirus tweets NLP - Text Classification \cite{miglani_2020}, and Trip Advisor Hotel Reviews \cite{ALAM2016206}. The Python code of the SADDA method for object recognition tasks can be downloaded at \url{https://github.com/NguyenThaiVu/SADDA}.

Our contributions can be summarized as follows:
\begin{itemize}
    \item We propose a new Semi-supervised Adversarial Discriminative Domain Adaptation method (SADDA) for addressing the unsupervised domain adaptation task.
    
    \item We illustrate that SADDA improves digit classification tasks and achieves competitive performance with other adversarial adaptation methods.
    
    \item We also demonstrate that the SADDA method can apply to multiple applications, including object recognition and natural language processing tasks.
    
\end{itemize}

% ====================================================================

\section{Related Work} \label{Sec:rw}

\textbf{Domain adaptation} is an active research field, which can handle numerous problems such as imbalanced data \cite{Krawczyk2016LearningFI}, dataset bias \cite{Torralba2011UnbiasedLA}, and domain shift \cite{chi2020collaborative}. Recent research has focused on domain adaptation from a labeled source dataset to an unlabeled target dataset, also known as unsupervised domain adaptation \cite{Kouw_2021,Margolis2011ALR}. The principle technique is minimizing the distinction between the source and target distribution \cite{Wang_2018}. Some popular approaches are Maximum Mean Discrepancy (MMD) \cite{5376}, deep reconstruction classification network (DRCN) \cite{ghifary2016deep} or Autoencoder-based domain adaptation \cite{6817520}.

\textbf{Adversarial-based domain adaptation}. With the rise of generative adversarial networks \cite{NIPS2014_5ca3e9b1}, the adversarial-based made huge advancements in the domain adaptation task \cite{long2017deep,long2018conditional,Saito_2018_CVPR}. Adversarial-based techniques try to achieve domain adaptation by using domain discriminators, which increases domain confusion through an adversarial process. A popular adversarial-based domain adaptation method is the Adversarial Discriminative Domain Adaptation (ADDA) by Tzeng \emph{et al.} \cite{8099799}. ADDA approach aims to diminish the distance between the source encoder and target encoder distributions through the domain-adversarial process. However, this method only distinguishes between the source and target domain. Instead, our SADDA method not only predicts whether the source domain or the target domain, but also classifies the label of the source dataset. More concretely, we force the adversarial-based method to the semi-supervised context. We will show that this creation can produce a more efficient classification model.

\textbf{Combining adversarial-based domain adaptation with other auxiliary tasks}. Recently, some works have focused on combining auxiliary tasks for adversarial-based adaptation to exploit more information \cite{ghifary2015domain,bousmalis2016domain}. Xavier and Bengio introduce Stacked Denoising Autoencoders \cite{DBLP:conf/icml/GlorotBB11,Vincent2010StackedDA}, reconstructing the merging data from numerous domains with the same network, such that the representations can be symbolized by both the source and target domain. Deep reconstruction classification network (DRCN) \cite{ghifary2016deep} attempts to solve two sub-problem at the same time: classification of the source data, and reconstruction of the unlabeled target data. However, these auxiliary tasks are not towards the same goal. We observe that during the adversarial process, we can classify the source or target dataset and predict the label of the source dataset simultaneously. That allows us to re-use the same output layers in the discriminator model as well as forces two discriminator models towards the same goal (sub-section \ref{Sec:discriminator} for more details). In addition, we also demonstrate that our SADDA method not only applies to computer vision tasks but also the natural language processing task.

% ==========================================================================
\section{Proposed method} 
\label{Sec:sadda}

% -------------------------------------------------

\subsection{\textbf{Semi-supervised Adversarial Discriminative Domain Adaptation}}

\begin{figure}[h]
    \includegraphics[width=0.99\columnwidth]{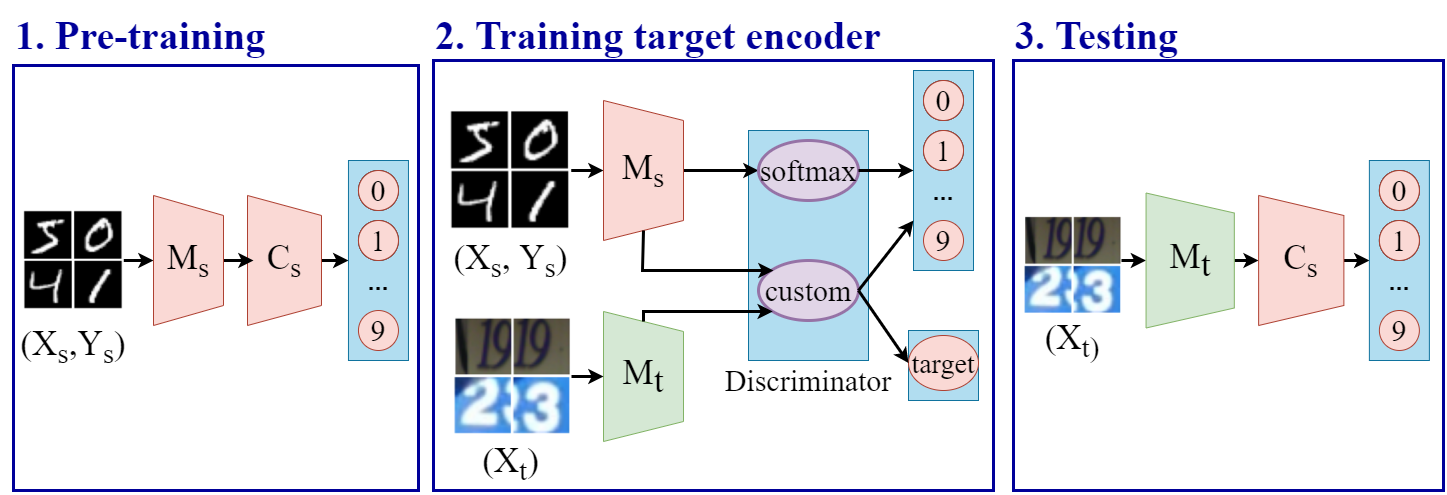}
    \caption{An overview of the SADDA. Firstly, training the source encoder ($M_s$) and the classification ($C_s$) using the source labeled images ($X_s$, $Y_s$). Secondly, training a target encoder ($M_t$) through the domain adversarial process. Finally, in the testing phase, concatenate the target encoder ($M_t$) and the classification ($C_s$) to create the complete model, which will predict the label of the target dataset precisely.}
    \label{fig:semi-adda}
\end{figure}

In this section, we describe in detail our Semi-supervised Adversarial Discriminative Domain Adaptation (SADDA) method. An overview of our method can be found in Figure \ref{fig:semi-adda}. 

In the unsupervised domain adaptation task, we already have source images $\mathbf{X}_s$ and source labels $\mathbf{Y}_s$ come from the source domain distribution $\mathbf{p}_{s}(x, y)$. Besides that, a target dataset $\mathbf{X}_t$ comes from a target distribution $\mathbf{p}_{t}(x, y)$, where the label of the target dataset is non-exist. We desire to learn a target encoder $M_t$ and classifier $C_t$, which can accurately predict the target image's label. In an adversarial-based adaptation approach, we aim to diminish the distance between the source mapping distribution ($M_{s}(X_s)$) and target mapping distributions ($M_{t}(X_t)$). As a result, we can straightly apply the source classifier $C_s$ to classify the target images, in other words, C = $C_t$ = $C_s$. The summary process of SADDA includes three steps: pre-training, training target encoder, and testing.

\textbf{Pre-training}. In the pre-training phase, training source encoder ($M_s$) and source classifier ($C_s$), by using the source labeled images ($\mathbf{X}_s$,$\mathbf{Y}_s$). This step is a standard supervised classification task, a common form can be denoted as:

\begin{equation}\label{supervised_loss}
    \argmin_{M_s, C_s} \mathcal{L}_{cls}(\mathbf{X}_s, \mathbf{Y}_s) = - \mathbb{E}_{(x,y){\sim}       (\mathbf{X}_s, \mathbf{Y}_s)} 
    \sum_{n=1}^{N} y_{_{n}} \log C_s(M_s(x_{_n}))
\end{equation}

where $\mathcal{L}_{cls}$ is a supervised classification loss (categorical crossentropy loss), and N is the number of classes.

\textbf{Training target encoder}. In the training target encoder phase, we first present a training discriminator process and then present a procedure for training the target encoder.

Firstly, training the discriminator (D) in two modes, each giving a corresponding output. (1) Supervised mode, where the supervised discriminator ($D_{sup}$) predicts N labels from the original classification task. (2) Unsupervised mode, where the unsupervised discriminator ($D_{unsup}$) classifies between $\mathbf{X}_s$ and $\mathbf{X}_t$. Discriminator correlates with unconstrained optimization:

\begin{equation}\label{supervised_discriminator_loss}
    \argmin_{D_{sup}} \mathcal{L}_{cls}
    (\mathbf{X}_s, \mathbf{Y}_s) = - \mathbb{E}_{(x,y){\sim} (\mathbf{X}_s, \mathbf{Y}_s)} 
    \sum_{n=1}^{N} y_{_{n}} 
    \log D_{sup}(M_s(x_{_n}))
\end{equation}

\begin{equation}
\label{unsupervised_discriminator_loss}
\begin{split}
    \argmin_{D_{unsup}} \mathcal{L}_{adv_D} 
    (\mathbf{X}_s, \mathbf{X}_t, M_s, M_t) = 
    - \mathbb{E}_{x_s {\sim} \mathbf{X}_s} 
    \log D_{unsup}(M_s(x_s))  \\
    - \mathbb{E}_{x_t {\sim} \mathbf{X}_t} 
    \log (1 - D_{unsup}(M_t(x_t)))
\end{split}
\end{equation} 

In equation \ref{supervised_discriminator_loss}, $\mathcal{L}_{cls}$ is a supervised classification loss corresponding to predicting N labels from the original classification task in the source dataset ($\mathbf{X}_s$), which will update the parameter in $D_{sup}$. In equation (\ref{unsupervised_discriminator_loss}), $\mathcal{L}_{adv_D}$ is an adversarial loss for unsupervised discriminator $D_{unsup}$, which trains ($D_{unsup}$) to maximize the probability of predicting the correct label from the source dataset or target dataset. One thing to notice is that the unsupervised discriminator uses a custom activation function (equation \ref{Custom_loss}), which returns a probability to determine whether a source image or target image (sub-section \ref{Sec:discriminator} for more details). 

Secondly, training the target encoder $M_t$ with the standard loss function and inverted labels \cite{NIPS2014_5ca3e9b1}. This implies that the unsupervised discriminator $D_{unsup}$ is fooled by the target encoder $M_t$, in other words, $D_{unsup}$ is unable to determine between $\mathbf{X}_s$ and $\mathbf{X}_t$. The feedback from the unsupervised discriminator $D_{unsup}$ allows the $M_t$ to learn how to produce a more authentic encoder. The loss for $\mathcal{L}_{{adv}_M}$ can be denoted:

\begin{equation} \label{target_encoder_loss}
    \argmin_{M_t} \mathcal{L}_{{adv}_M} 
    (\mathbf{X}_s, \mathbf{X}_t, D) = 
    - \mathbb{E}_{x_t {\sim} X_t} \log  D_{unsup}(M_t(x_t))
\end{equation}

\textbf{Testing}. In the testing phase, we concatenate the target encoder $M_t$ and the source classifier $C_s$ to predict the label of target images $\mathbf{X}_t$. 

% ---------------------------------------------------------------
\subsection{\bf The discriminator model}
\label{Sec:discriminator}

In this section, we describe the detail of the discriminator model and provide some arguments to prove the effectiveness of the discriminator model in the SADDA method.

In the training target encoder step, the discriminator model is trained to predict N+1 classes, where N is the number of classes in the original classification task (supervised mode) and the final class label predicts whether the sample comes from the source dataset or target dataset (unsupervised mode). The supervised discriminator and the unsupervised discriminator have different output layers but have the same feature extraction layers - via backpropagation when we train the network, updating the weights in one model will impact the other model as well.

The supervised discriminator model produces N output classes (with a softmax activation function). The unsupervised discriminator is defined such that it grabs the output layer of the supervised mode \textit{prior softmax activation} and computes a normalized sum of the exponential outputs (custom activation). When training the unsupervised discriminator, the source sample will have a class label of 1.0, while the target sample will be labeled as 0.0. The explicit formula of custom activation \cite{salimans2016improved} is: 

\begin{equation}\label{Custom_loss}
    D(x) = \frac{Z(x)}{Z(x) + 1}  
\end{equation}
where
\begin{equation}
Z(x) = \sum_{n=1}^{N} \exp[l_n(x)]
\end{equation}

\begin{table}
\caption{Experimental compute on custom activation - the output of unsupervised discriminator model}
\label{experimental_custom}  
\begin{tabular}{llll}
\hline
Output probabilities (prior softmax) & Custom activation & Entropy  \\
\hline
{[9.0, 1.0, 1.0]} & 0.9999 & Low \\
{[5.0, 1.0, 1.0]} & 0.9935 & Low \\
{[-5.0, -5.0, -5.0]} & 0.0198 & High  \\
\hline
\end{tabular}
\end{table}

The experiment of equation (\ref{Custom_loss}) is described in Table \ref{experimental_custom}, and the outputs are between $0.0$ and $1.0$. If the probability of output value \textit{prior softmax activation} is a large number (meaning: low entropy) then the custom activation output value is close to 1.0. In contrast, if the output probability is a small value (meaning: high entropy), then the custom activation output value is close to 0.0. Implied that the discriminator is encouraged to output a confidence class prediction for the source sample, while it predicts a small probability for the target sample. That is an elegant method allowing re-use of the same feature extraction layers for both the supervised discriminator and the unsupervised discriminator.

It is reasonable that learning the well supervised discriminator will improve the unsupervised discriminator. Moreover, training the discriminator in unsupervised mode allows the model to learn useful feature extraction capabilities from huge unlabeled datasets. As a sequence, improving the supervised discriminator will improve the unsupervised discriminator and vice versa. Improving the discriminator will enhance the target encoder  \cite{odena2016semisupervised}. In total, this is one kind of advantage circle, in which three elements (unsupervised discriminator, supervised discriminator, and target encoder) iteratively make each other better. 

% -----------------------------------------------------------
\subsection{\bf Guideline for stable SADDA}
\label{stable_SADDA}

In general, training SADDA is an extremely hard process, there are two losses we need to optimize: the loss for the discriminator and the loss for the target encoder. For that reason, the loss landscape of SADDA is fluctuating and dynamic (detail in sub-section \ref{convergence_analysis}). When implementing and training the SADDA, we find that is a tough process. To overcome the limitation of the adversarial process, we present a full architecture of SADDA. This designed architecture increases training stability and prevents non-convergence. In this section, we present the key ideas in designing the model for the image classification and the sentiment classification task.

\textbf{Image classification}. The design of SADDA is inspired by Deep Convolutional GAN (DCGAN) architecture \cite{radford2016unsupervised}. The summary architecture of the SADDA method for digit recognition is shown in Figure \ref{fig:sadda-digit}. On the one hand, the encoder is used to capture the content in the image, increasing the number of filters while decreasing the spatial dimension by the convolutional layer. On the other hand, the discriminator is symmetric expansion with the encoder by using fractionally-strided convolutions (transpose convolution).

Moreover, our recommendation for efficient training SADDA in the image classification task: 
\begin{itemize}
    \item[$\bullet$] Replace any pooling layers with convolution layers (or transposed convolution) with strides larger than 1. 
    \item[$\bullet$] Remove fully connected layers in both encoder and discriminator (except the last fully connected layers, which are used for prediction).
    \item[$\bullet$] Use ReLU activation \cite{Maas13rectifiernonlinearities} in the encoder and LeakyReLU activation \cite{Maas13rectifiernonlinearities} (with alpha=0.2) in the discriminator. 
\end{itemize}

\textbf{Sentiment classification}. The design of the SADDA method for sentiment classification is inspired by the architecture called Autoencoders LSTM \cite{https://doi.org/10.48550/arxiv.1502.04681,brownlee_2020}. The summary architecture of the SADDA method for sentiment classification is demonstrated in Figure \ref{fig:sadda-nlp}. In general, the architecture of the model used in the sentiment classification task has many similarities with the architecture used in image classification. Firstly, we remove fully connected layers in both the encoder and the discriminator. Instead, we use the Long Short Term Memory \cite{article_lstm} (LSTM) to handle sequences of text data. Secondly, the network is organized into an architecture called the Encoder-Decoder LSTM, with the Encoder LSTM being the encoder block and the Decoder LSTM being the discriminator block respectively. The Encoder-Decoder LSTM was built for the NLP task where it illustrated state-of-the-art performance, such as machine translation \cite{cho-etal-2014-learning}. From the empirical, we find that the Encoder-Decoder is suitable for the unsupervised domain adaptation task.

% ==================================================================
\section{Experiments} 
\label{Sec:exp}

In this section, we evaluate our SADDA method for unsupervised domain adaptation tasks in three scenarios: digit recognition, object recognition, and sentiment classification. 

In the experiments, we focus on probing how the SADDA method improves the unsupervised domain adaptation task. For this purpose, we only choose shallow architecture rather than a deep network. We leave the sophisticated design for a future job.

% ------------------------------------------------------------------
\subsection{Digit recognition}

\subsubsection{Datasets and Domain Adaptation Scenarios}

We evaluate SADDA on various unsupervised domain adaptation experiments, examining the following popular used digits datasets and settings (the visualization is in Figure \ref{fig:samle_images}): 

\textbf{MNIST} $\longleftrightarrow$ \textbf{USPS}: MNIST \cite{726791} includes 28x28 pixels, which are grayscale images of digit numbers. USPS \cite{uspsdataset} is a digit dataset, which contains 9298 grayscale images. The image is 16x16 pixels. In this experiment, we follow the evaluation protocol of \cite{bousmalis2017unsupervised}.

\textbf{MNIST} $\rightarrow$ \textbf{MNIST-M}: MNIST-M \cite{ganin2016domainadversarial} is made by merging MNIST digits with the patches arbitrarily extracted from color images of BSDS500 \cite{5557884}. In this experiment, we set the input size is 28x28x3 pixels, and we follow the evaluation protocol of \cite{bousmalis2017unsupervised}

\textbf{SVHN} $\rightarrow$ \textbf{MNIST}: The Street View House Number (SVHN) \cite{37648} is a digit dataset, which contains 600000 32×32 RGB images. In this experiment, we convert the SVHN dataset to grayscale images and resize the MNIST images into 32x32 grayscale images. We use the evaluation protocol of \cite{ghifary2016deep}.

% -------------------------------------
\subsubsection{Implementation details}
\label{5_implement_digit}

The SADDA model is trained with different learning rates in different phases. In the pre-training phases, this is a standard classification task, we use a learning rate is 0.001 in our experiment. In the training target encoder phases, we suggested a learning rate of 0.0002 as well as an Adam optimizer \cite{kingma2017adam} and setting the $\beta_1$ equal to 0.5 to help stabilize training. In the LeakyReLU activation, we set $\alpha$ = 0.2 in the whole model.

In this experiment, the encoder consists of four convolutional layers with 4 x 4 kernel size, 2 x 2 strides, same padding, and ReLU activation. The number of filters for four convolution layers are 32, 64, 128, and 256, respectively. The target encoder has the same architecture as the source encoder, and the source encoder is used as an initialization for the target encoder. 

The classifier takes the outputs of the encoder as input. Next, a fully connected layer with 100 feature channels, followed by ReLU activation. Finally, the fully connected layer with ten feature channels and the softmax activation.

In the training target encoder phases, the outputs of the encoder serve as the input of the discriminator. The discriminator consists of four Transpose Convolutional layers with 4 x 4 kernels size, 2 x 2 strides, the same padding, and the Leaky ReLU activation (alpha = 0.2). The number of kernels for four Transpose Convolutional is 256, 128, 64, and 32, respectively. We illustrate the overall architecture in Figure \ref{fig:sadda-digit}.

\subsubsection{Results on digit datasets}
\label{sec:result_digit}

In this experiment, we compare our SADDA method against multiple state-of-the-art unsupervised domain adaptation methods.

\begin{table}[h]
\caption{Experimental results on unsupervised domain adaptation on digit datasets. The results are not re-implement, instead, we select based on the available result in the previous publication (some experimental results have the standard deviation because that publication has the standard deviation while others do not.)}
\label{tab_result_digit}   
\begin{tabular}{llllll}
\hline\noalign{\smallskip}
Method & mnist$\rightarrow$usps & usps$\rightarrow$mnist & mnist$\rightarrow$mnist-m & svhn$\rightarrow$mnist & \\

\noalign{\smallskip}\hline\noalign{\smallskip}

Source only & 78.9 & 57.1 \textpm 1.7 & 63.6 & 60.1 \textpm 1.1  \\

DANN \cite{ganin2016domainadversarial} & 85.1 & 73.0 \textpm 2.0 & 77.4 & 73.9 \\

DRCN \cite{ghifary2016deep} & 91.8 \textpm 0.1 & 73.7 \textpm 0.1 & - & 82.0 \textpm 0.2 \\

ADDA \cite{8099799} & 89.4 \textpm 0.2 & 90.1 \textpm 0.8 & - &  76.0 \textpm 1.8 \\

SBADA-GAN \cite{russo2017source} & 97.6 & 95.0 & \bf 99.4 & 76.1 \\

SHOT \cite{liang2020we} & 98.0 & \bf 98.4 & - & \bf 98.9 \\

DFA-MCD \cite{wang2021discriminative} & \bf 98.6 & 96.6 & - & \bf 98.9 \\

\hline 
SADDA (our) & 98.1 & 97.8 & 78.2 & 86.5 \\
\hline

\end{tabular}
\end{table}

Experimental results are shown in Table \ref{tab_result_digit}. In the real-world dataset SVHN $\rightarrow$ MNIST, the SADDA model showed approximately 26{\%} improvement over the Source only model, 10{\%} more than the ADDA method \cite{8099799}. In addition, in the first two experiments (MNIST $\rightarrow$ USPS and USPS $\rightarrow$ MNIST), the SADDA method achieves extremely high accuracy, which results in 98.1{\%} and 97.8{\%} respectively. However, SADDA has a little lower accuracy than other methods like SBADA-GAN \cite{russo2017source}, SHOT \cite{liang2020we}, and DFA-MCD \cite{wang2021discriminative} in some experiments.

Although, our method did not achieve the highest accuracy in any of the experiments. Our method still has competitive accuracy when compared with the state-of-the-art methods in the last two years (SHOT \cite{liang2020we}, DFA-MCD \cite{wang2021discriminative}). For example, our SADDA method compared with the DFA-MCD method, in the MNIST $\rightarrow$ USPS experiment, we have lower accuracy (98.1$\%$ versus 98.6$\%$); however, in the USPS $\rightarrow$ MNIST experiment, we achieved a higher accuracy (SADDA achieved 97.8$\%$ compared to 96.6$\%$).

\begin{figure}[h]
    \includegraphics[width=0.95\columnwidth]{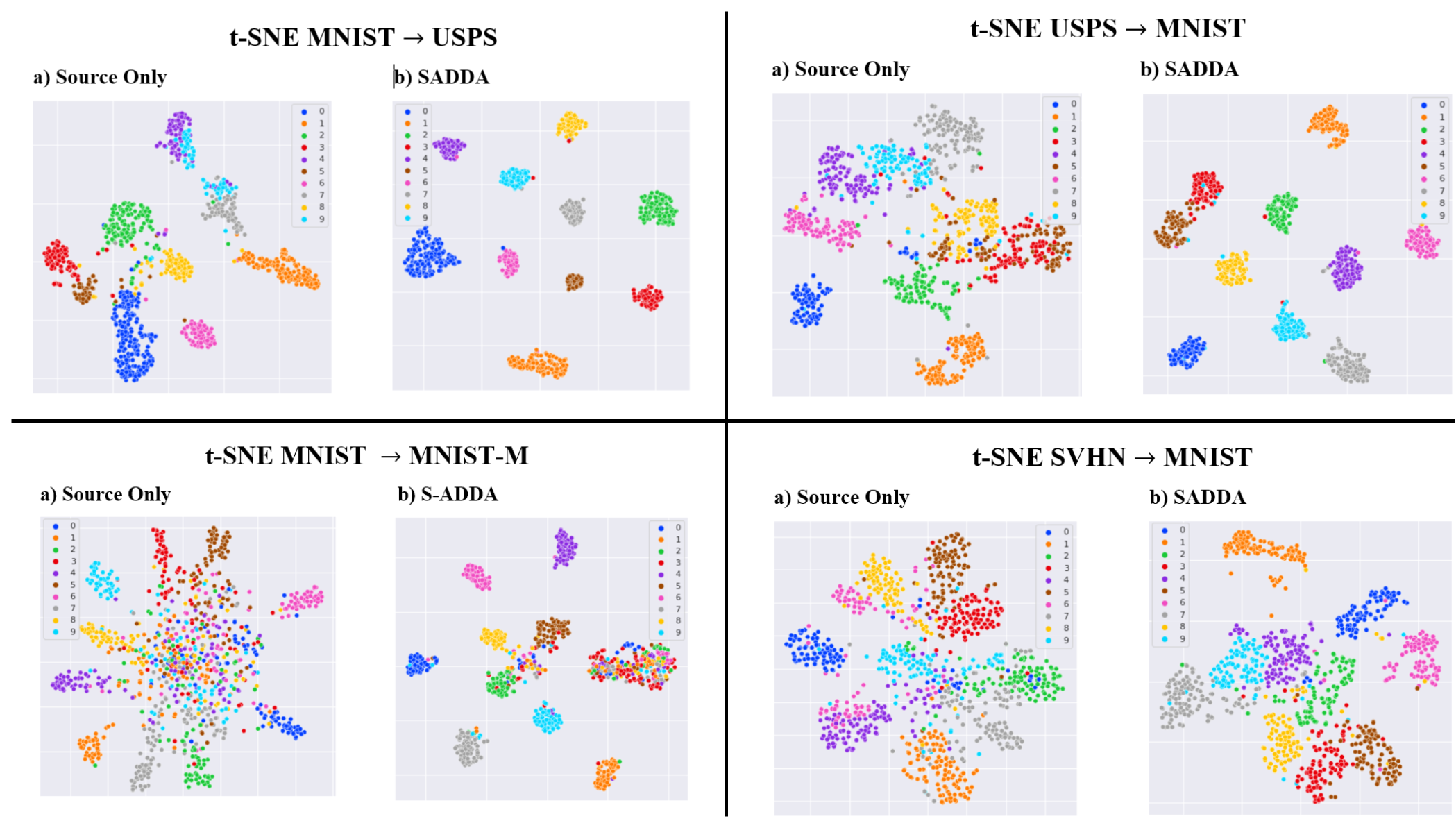}
    \caption{t-SNE embedding of digit classification, using (2 x 2 x 256) dimensional representation, with Source only (on the left) and SADDA (on the right) on the target dataset. Note that SADDA minimizes intra-class distance and maximizes inter-class distance.}
    \label{fig:tSNE_SADDA}
\end{figure}

For further insight into the SADDA model effect on the digit classification tasks, we use t-SNE \cite{JMLR:v9:vandermaaten08a} to visualize the 2D point of the last encoder layer of SADDA (as described in Figure \ref{fig:tSNE_SADDA}). Ten labels are from 0 to 9 corresponding, and 100 samples per label. The domain invariance is determined by the degree of overlap between features. Regarding the Source only model, the distribution and the density are messy. In contrast, the SADDA method splits different labels into different regions, and the overlap is more prominent.

% ------------------------------------------------------------------

\subsection{Object recognition}

\subsubsection{Datasets and Preprocessing}

In this subsection, we present the experiments for evaluating the SADDA method. The experiment is performed on the VLCS \cite{Torralba2011UnbiasedLA} dataset, including PASCAL VOC2007 (V) \cite{pascal-voc-2007}, LABELME (L) \cite{Russell2007LabelMeAD}, CALTECH (C) \cite{1597116}, and SUN (S) \cite{5540221} datasets. Each dataset contains five categories: bird, car, chair, dog, and person. Since the number of images per class is not equal, we use data augmentation techniques to balance the number of images. In this experiment, we use the Albumentations library \cite{info11020125} to increase to 5000 images per class. We divide the dataset into a training set (60$\%$), validation set (20$\%$), and test set (20$\%$).

The detailed architecture is shown in Figure \ref{fig:sadda-object}. The rest of the other installation (optimization algorithm, learning rate) is the same as section \ref{5_implement_digit}.

In the experiments, one dataset is used as the source domain and the rest is used as the target domain, resulting in four different cases (Table \ref{tab:object}). In addition, we do not compare our SADDA model with other domain adaptation methods due to different setups.

\subsubsection{Results}

The results on the VLCS are shown in Table \ref{tab:object}. Source only is a model that only trains on the source dataset without using any domain adaptation methods. Overall, the accuracy when applying the SADDA method overcomes the Source only model in all cases. In some specific cases like PASCAL $\rightarrow$ CALTECH, the classification accuracy goes from 45.10 to 55.30 (improving approximately 10$\%$). In case LABELME $\rightarrow$ CALTECH, the accuracy grows from 32.75$\%$ to 39.36$\%$. 

Examining the results in Table \ref{tab:object}, the Source only model has low accuracy, which reveals that the domain shift is quite large. In other words, the Source only model does not learn any knowledge about the source dataset to predict the target dataset. In contrast, the SADDA model learns a more useful feature representation, leading to higher accuracy when performing a prediction on the target dataset.

Although the SADDA method has certain improvements compared to the Source only method, the accuracy of the SADDA method is still low and not ideal. For further comparison, we also test the hypothesis situation where the target labels are present (the train on target model). There is still a big gap between the accuracy of the SADDA method and the train on target method.

\begin{table}

    \begin{subtable}[h]{0.7\columnwidth}
        \centering
        \begin{tabular}{llll}
        & LABELME & CALTECH & SUN \\
        \hline
        Source only & 33.26 & 45.10 & 33.78 \\
        SADDA & 37.73 & 55.30 & 36.21 \\
        Train on target & 86.52 & 99.27 & 88.26\\
       \end{tabular}
       \caption{Source domain: PASCAL}
       \label{tab:pascal}
    \end{subtable}
    
    \hfill
    
    \begin{subtable}[h]{0.7\columnwidth}
        \centering
        \begin{tabular}{llll}
        & PASCAL & CALTECH & SUN \\
        \hline 
        Source only & 28.73 & 32.75 & 27.71 \\
        SADDA & 32.71 & 39.36 & 31.01 \\
        Train on target & 75.28 & 99.27 & 88.26 \\
        \end{tabular}
        \caption{Source domain: LABELME}
        \label{tab:labelme}
     \end{subtable}
     
     \begin{subtable}[h]{0.7\columnwidth}
        \centering
        \begin{tabular}{llll}
        & PASCAL & LABELME & SUN \\
        \hline 
        Source only & 26.71 & 28.27 & 31.62 \\
        SADDA & 31.82 & 31.39 & 37.67 \\
        Train on target & 75.28 & 86.52 & 88.26 \\
       \end{tabular}
       \caption{Source domain: CALTECH}
       \label{tab:caltech}
    \end{subtable}
    
    \hfill
     
     \begin{subtable}[h]{0.7\columnwidth}
        \centering
        \begin{tabular}{llll}
        & PASCAL & LABELME & CALTECH \\
        \hline
        Source only & 29.72 & 25.87 & 39.66 \\
        SADDA & 33.04 & 27.35 & 41.52 \\
        Train on target & 75.28 & 86.52 & 99.27 \\
       \end{tabular}
       \caption{Source domain: SUN}
       \label{tab:sun}
    \end{subtable}
    
    \hfill
     
     \caption{The accuracy ($\%$) on the VLCS dataset.}
     \label{tab:object}
\end{table}

% ------------------------------------------------------------------
\subsection{Sentiment classification}

\subsubsection{Datasets and Preprocessing}

In this subsection, we evaluate the SADDA method for the sentiment classification task. We use three sentimental datasets, including Women's E-Commerce Clothing Reviews \cite{nicapotato_2018}, Coronavirus tweets NLP - Text Classification \cite{miglani_2020}, and Trip Advisor Hotel Reviews \cite{ALAM2016206}:

\textbf{Women's E-Commerce Clothing Reviews} \cite{nicapotato_2018}. This is real commercial data, where the reviews are written by customers. In this task, we only use two features called \textit{Review Text} (the raw text review) and \textit{Rating} (the positive integer for the product, provided by the customer from 1 Worst to 5 Best). Regarding the \textit{Rating}, we relabel into the sets \{positive, neutral, negative\} with the following rule: if a review is greater than 3, it is considered a positive comment; if a review is equal to 3, it is considered a neutral comment; if a review is less than 3, it is considered a negative comment.

\textbf{Coronavirus tweets NLP - Text Classification} \cite{miglani_2020}. The tweets were downloaded from Twitter and tagged manually. Although there are four columns in total, we only use the \textit{Original Tweet} feature and \textit{Label} in our experiment. In the case of \textit{Label}, the original label includes {Extremely Negative, Negative, Neutral, Positive, and Extremely Positive}. However, we convert to the sets \{positive, neutral, negative\} respectively.

\textbf{Trip Advisor Hotel Reviews} \cite{ALAM2016206}. This contains reviews crawled from the travel company called Tripadvisor. The dataset contains two features, including \textit{Review Text} and \textit{Rating} (the positive integer from 1 Worst to 5 Best). Regarding \textit{Rating}, we process the same as the case Women's E-Commerce Clothing Reviews above.

In all three datasets, we perform the following text preprocessing steps: removing the punctuation, URL, hashtags, mentions, and stop words (with the support of the NLTK \cite{nltk_library} library). We limit the input sentence to a max length equal to 50. The GloVe \cite{pennington-etal-2014-glove} word embedding is applied to map the word in the text review to the vector space. 

Because the number of samples per class is not balanced, we use the data augmentation techniques to balance the number of samples - with the support of the TextAugment \cite{marivate2020improving} library. Particularly, we perform data augmentation such that each class has up to 20 000 samples. The dataset is divided into a training set (60$\%$), validation set (20$\%$), and test set (20$\%$).

\subsubsection{Experiments and Results}

For this experiment, our architecture is illustrated in Figure \ref{fig:sadda-digit}. Elements in that architecture such as the LSTM layer and the Repeat Vector layer are implemented by the TensorFlow library with default settings. The optimization algorithms and learning rates are set up as in section \ref{5_implement_digit}. In addition, we do not attempt to fine-tune the architecture and leave it for future work.

The results of our experiment are provided in Table \ref{table_result_nlp}. Compared with the Source only model, the SADDA method shows a little improvement in the accuracy of this sentiment analysis task. For a certain experiment, like T $\rightarrow$ W, the classification accuracy goes from 41.07$\%$ to 49.28$\%$. However, not all experiments improve, such as experiment T $\rightarrow$ C, the accuracy even dropped a bit (from 38.46$\%$ down to 38.05$\%$). Additionally, a comparison with the "Train on target" model exposes that the SADDA model is far from the ideal model. We hope that is the motivation for future development. 

\begin{table}[h]
\caption{The accuracy ($\%$) of unsupervised domain adaptation on the sentiment classification task. In the table, there are three datasets: Women's E-Commerce Clothing Reviews (W) \cite{nicapotato_2018}, Coronavirus tweets (C) \cite{miglani_2020}, Trip Advisor Hotel Reviews (T) \cite{ALAM2016206}}
\label{table_result_nlp}   
\begin{tabular}{llllllll}
\hline\noalign{\smallskip}

& W $\rightarrow$ C & W $\rightarrow$ T & C $\rightarrow$ W & C $\rightarrow$ T & T $\rightarrow$ W & T $\rightarrow$ C & \\
\noalign{\smallskip}\hline
 
Source only & 37.97 & 50.11 & 45.91 & 49.93 & 41.07 & 38.46  \\

SADDA & 43.88 & 55.54 & 48.02 & 56.10 & 49.28 & 38.05 \\

\noalign{\smallskip}\hline

Train on target & 79.01 & 96.10 & 93.02 & 96.10 & 93.02 & 79.01 \\

\noalign{\smallskip}\hline
\end{tabular}
\end{table}

% ------------------------------------------------------------------
\subsection{Challenge and convergence analysis}
\label{convergence_analysis}

\hl{In this subsection, we will discuss the challenge of training a stable SADDA model and how to trigger the early stopping of the training progress to achieve the convergent state.

In the SADDA model, we have three stages: Pre-training, Training target encoder, and Testing. The difficulty comes from the Training target encoder process. The reason is that both the discriminator and target encoder are trained simultaneously in that procedure, which can lead to updating the parameter of one model will reduce the performance of the other model. More concretely, there are two loss functions we need to optimize: the discriminator loss and the adversarial loss. The discriminator loss (equation \ref{unsupervised_discriminator_loss}) is the loss when the unsupervised discriminator predicts the source or target samples. The adversarial loss (equation \ref{target_encoder_loss}) is the loss of the target encoder when training with the inverted labels.

\begin{figure} [htp]
    \includegraphics[width=0.75\columnwidth]{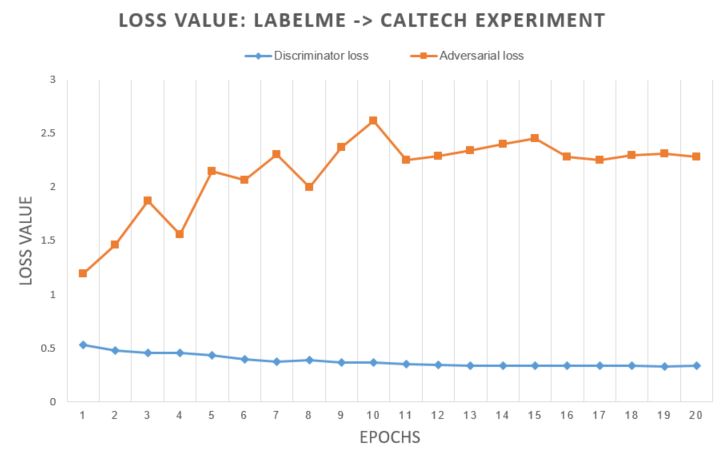}
    \caption{The discriminator loss and adversarial loss in the LABEL $\rightarrow$ CALTECH experiment.}
    \label{fig:losses_landscape}
\end{figure}

When training the target encoder, we do not try to find a minimum value for either discriminator loss or adversarial loss. Instead, we are looking for a Nash equilibrium state for both losses \cite{salimans2016improved}. In practice, we observe the discriminator loss and adversarial loss after each epoch, when both loss values no longer change, we trigger an early stopping. Early stopping is the technique to stop the training process at a certain point before the model overfits the training dataset and has poor performance on the test set. In that case, we consider that our model has converged. Keep in mind that the loss of 0.0 in the discriminator loss or the adversarial loss during the training process is a failure mode.

Regarding figure \ref{fig:losses_landscape}, the convergence point is in the epoch 20. At that point, we will stop the training process because the discriminator loss and the adversarial loss are saturated at around 0.33 and 2.30 respectively. In that experiment (and other object recognition tasks), it took around 1 hour on a single Tesla T4 GPU to complete the training procedure.}

% ====================================================================
\section{Conclusion}

We proposed a more stable and high-accuracy architecture for training adversarial-based domain adaptation methods. The key idea of this approach is to train discriminators in two modes: supervised mode and unsupervised mode. Moreover, utilize this to create a more efficient target encoder, which will help improve the classification accuracy. 

While the SADDA method has demonstrated an improvement in many tasks like image classification or sentiment classification, there are still open challenges. Particularly, the SADDA model in object recognition and sentiment classification is far from the desired accuracy model. We hope that the intuition of this research will facilitate further advances in domain adaptation tasks.


\bibliographystyle{ieeetr}  % mathematics and physical sciences
\bibliography{reference}

% ===============================================================
\begin{appendices}
In this appendix section, we present in detail the design architecture of our SADDA method in three experiments, including digit recognition, object recognition, and sentiment classification.

\begin{figure}[h]
    \includegraphics[width=0.9\columnwidth]{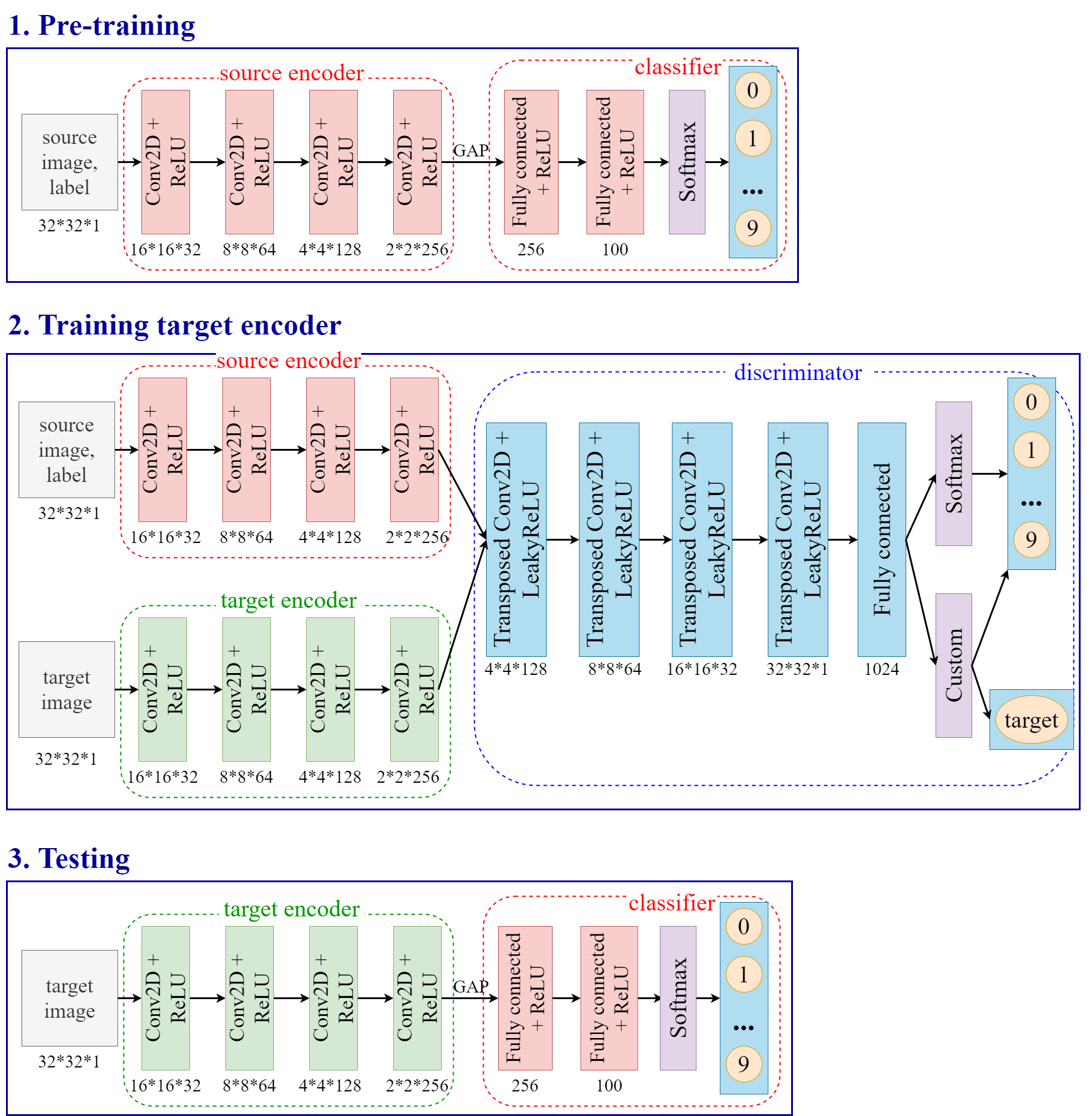}
    \caption{The overview of the SADDA method for the digit recognition task.  We found that Global Average Pooling (GAP) \cite{lin2014network} increased model stability and reduce the number of parameter.}
    \label{fig:sadda-digit}
\end{figure}

\begin{figure}[h]
    \includegraphics[width=0.9\columnwidth]{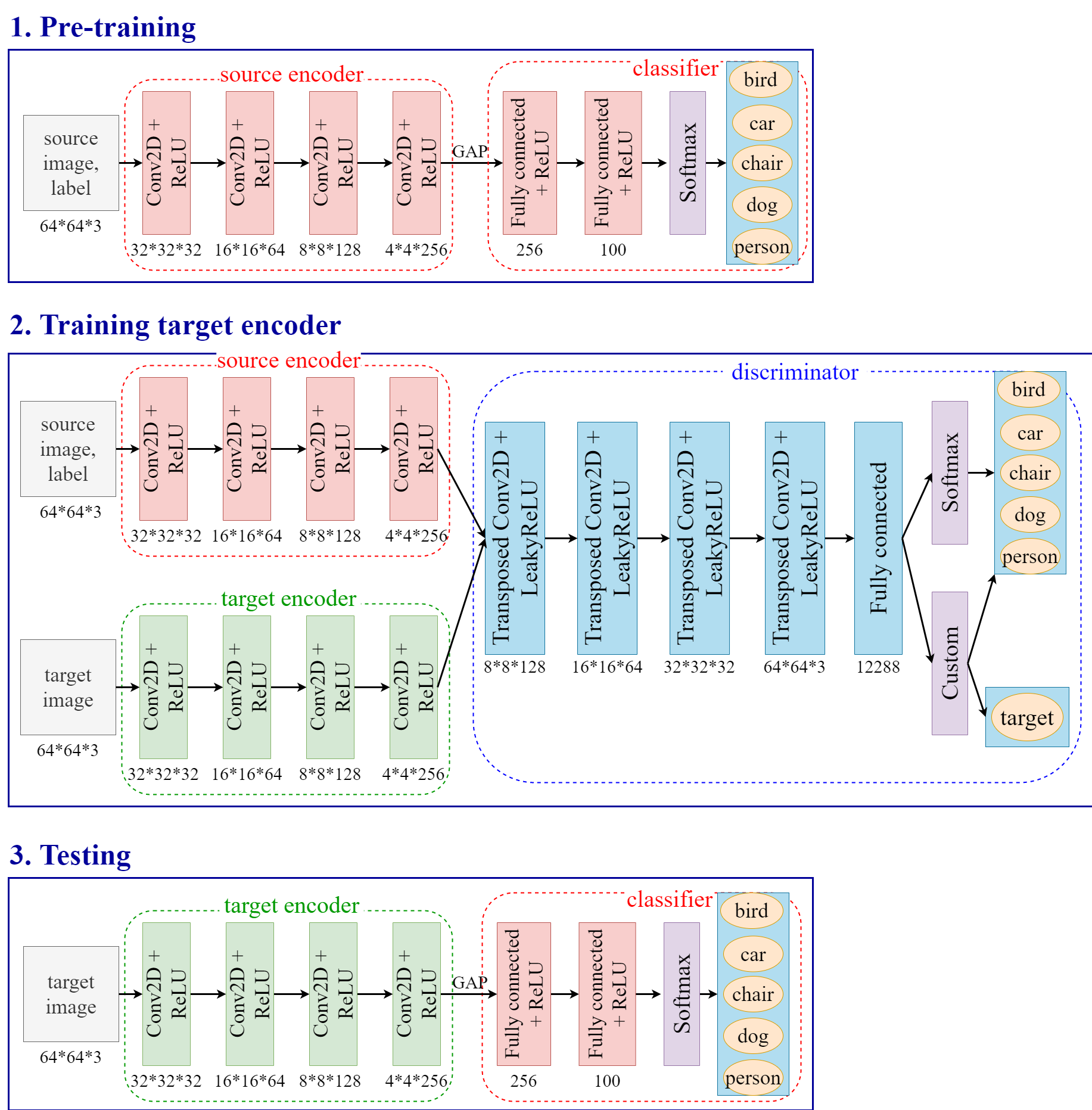}
    \caption{The overview of the SADDA method for the object recognition task on the VLCS \cite{Torralba2011UnbiasedLA} dataset. With the input image's shape is 64 x 64 x 3}
    \label{fig:sadda-object}
\end{figure}

\begin{figure}[h]
    \includegraphics[width=0.9\columnwidth]{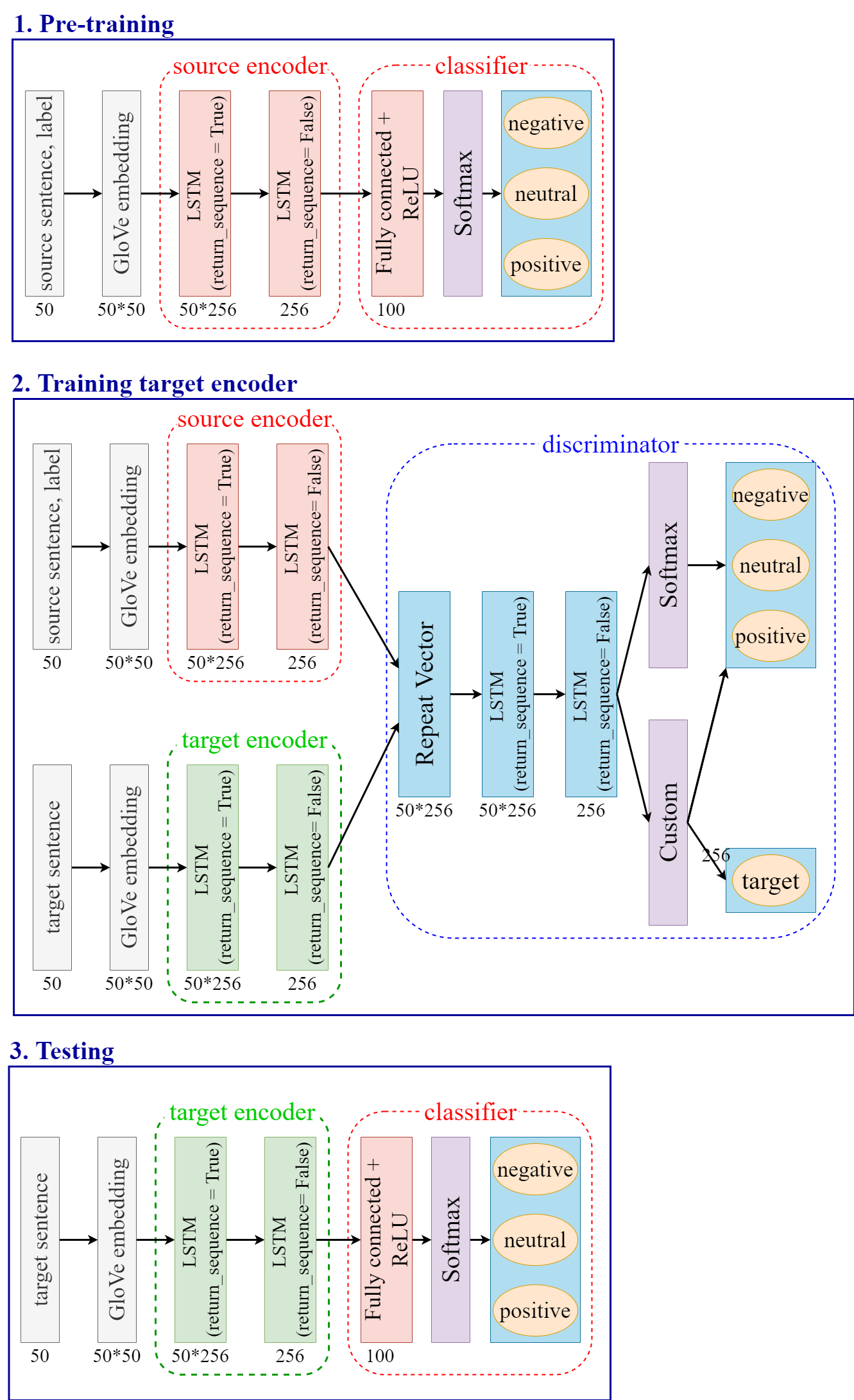}
    \caption{The overview of the SADDA method for the sentiment classification task. The input sentence has a max length equal to 50. In the design above, to prevent overfitting, the LSTM layer is always followed by the dropout layer with 0.2 rates. The numbers under the particular layer are the output shape of that layer.}
    \label{fig:sadda-nlp}
\end{figure}

\end{appendices}

\end{document}